\documentclass[runningheads]{llncs}

\usepackage[final,year=2026]{accv}

\usepackage{accvabbrv}
\usepackage{graphicx}
\usepackage{array}     % >{...}m{...} columns for the model-comparison grid
\usepackage{booktabs}
\usepackage{makecell}   % provides \Xhline used by the result tables
\usepackage{amsmath}
\usepackage{amssymb}
\usepackage{tabularx}
\usepackage{subcaption}
\usepackage[accsupp]{axessibility}
\AtBeginDocument{\hypersetup{
  hidelinks,
  pdftitle={DART: A Degradation-Aware Recurrent Transformer for Archival Film Restoration},
  pdfauthor={Mikolaj Jastrzebski, Wojciech Kozlowski, Kamil Adamczewski},
  pdfsubject={Archival film restoration},
  pdfkeywords={Old film restoration, Video restoration, Degradation-aware, Recurrent transformer, Mask supervision}
}}

\usepackage[breaklinks,colorlinks,citecolor=accvblue]{hyperref}
\usepackage{orcidlink}

\graphicspath{{figures/}}

\newlength{\modelcmppad}
\newlength{\modelcmplabelwidth}
\newlength{\modelcmpimgwidth}
\newlength{\modelcmpgridheight}
\newcommand{\modelcmpgrid}[1]{%
  \sbox0{#1}%
  \dimen1=\ht0\advance\dimen1\dp0%
  \ifdim\dimen1>\modelcmpgridheight
    \resizebox{!}{\modelcmpgridheight}{#1}%
  \else
    \resizebox{\linewidth}{!}{#1}%
  \fi
}
\newcommand{\modelcmptemporalrow}[4]{%
  \rotatebox{90}{\footnotesize#1} &
  \includegraphics[width=\linewidth]{#2} &
  \includegraphics[width=\linewidth]{#3} &
  \includegraphics[width=\linewidth]{#4} \\[3pt]
}
\let\modelcmpthreerow\modelcmptemporalrow

\begin{document}

% ---------------------------------------------------------------
\title{DART: A Degradation-Aware Recurrent Transformer for Archival Film Restoration}
\titlerunning{DART: Degradation-Aware Recurrent Transformer}

% Camera-ready: list authors and affiliations.
\author{Miko\l{}aj Jastrz\k{e}bski\orcidlink{0009-0003-3526-935X} \and
Wojciech Koz\l{}owski\orcidlink{0009-0009-1532-4415} \and
Kamil Adamczewski\orcidlink{0000-0002-2917-4392}}
\authorrunning{M.~Jastrz\k{e}bski et al.}
\institute{Wroc\l{}aw University of Science and Technology, Wroc\l{}aw, Poland\\
\email{mikjas02@gmail.com, \{wojciech.kozlowski,kamil.adamczewski\}@pwr.edu.pl}}

\maketitle

% --- ENTRY / TEASER figure on page 1, under the title block and above the
%     abstract (BasicVSR++ / BOF style). [H] pins it here. To be produced. ---
{%
  % Tighten vertical whitespace above/below this pinned [H] figure only.
  \setlength{\intextsep}{6pt}% space above/below in-text floats ([H])
  \captionsetup{skip=2pt}% space between figure content and its caption
  \begin{figure}[htbp]
    \centering
    \begin{subfigure}[b]{0.61\linewidth}
      \centering
      % No horizontal gaps between the three example frames.
      \setlength{\tabcolsep}{0pt}%
      \renewcommand{\arraystretch}{1}%
      % Force the grid to fit exactly inside this subfigure width.
      \resizebox{\linewidth}{!}{%
        \begin{tabular}{@{}c@{\hspace{2pt}}ccc@{}}
            \includegraphics{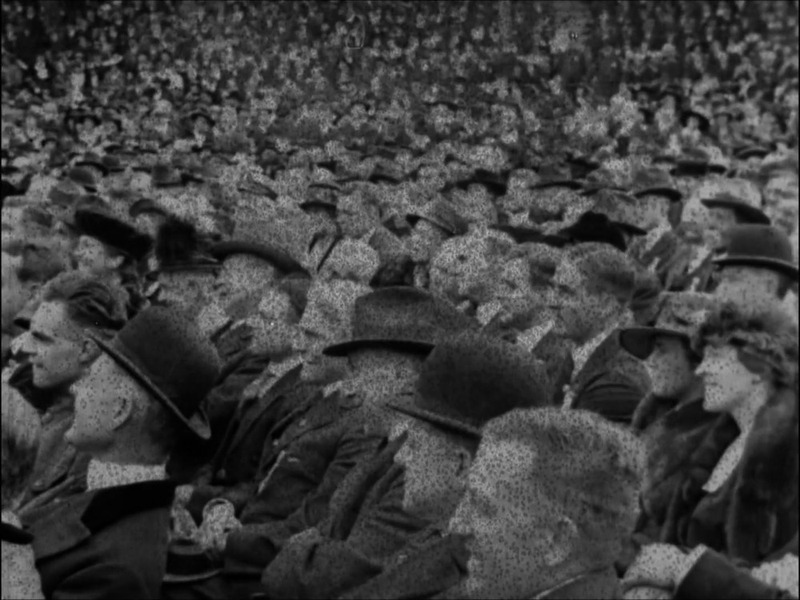}&
            \includegraphics{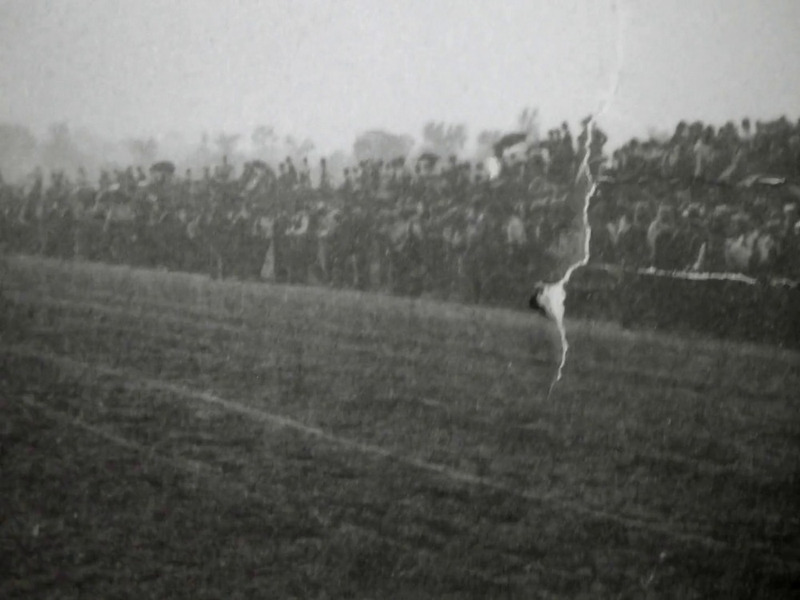}&
            \includegraphics{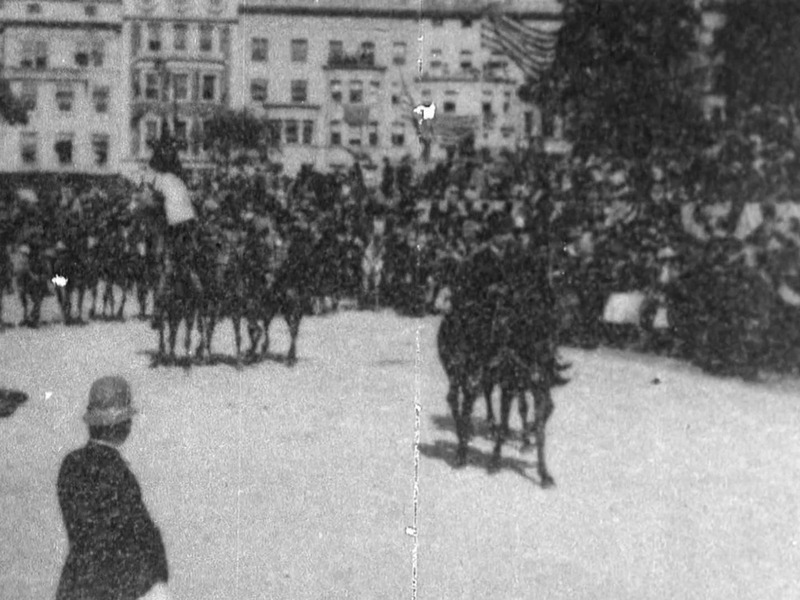} \\[-2.2pt]
            \includegraphics{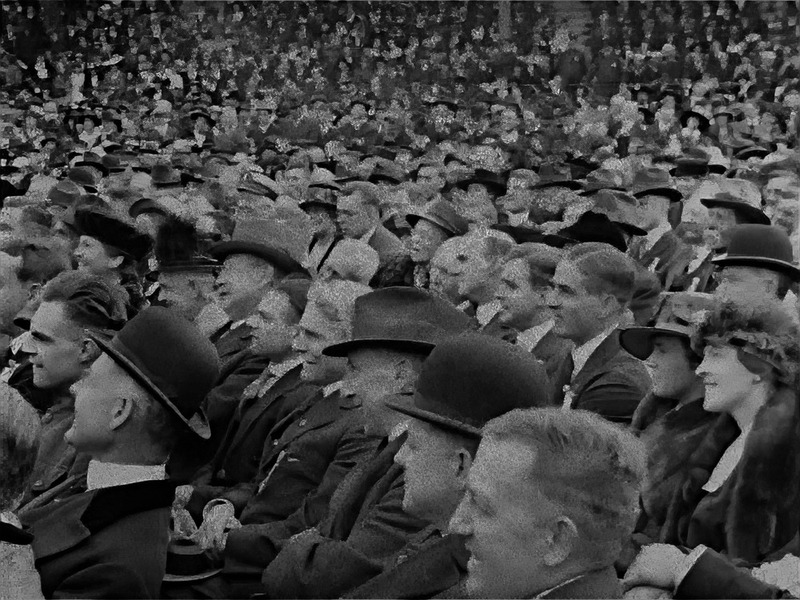}&
            \includegraphics{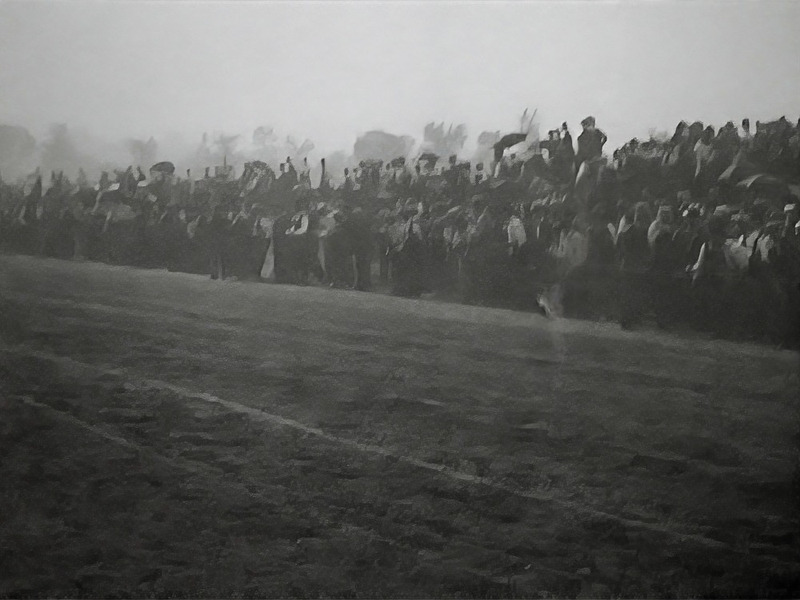}&
            \includegraphics{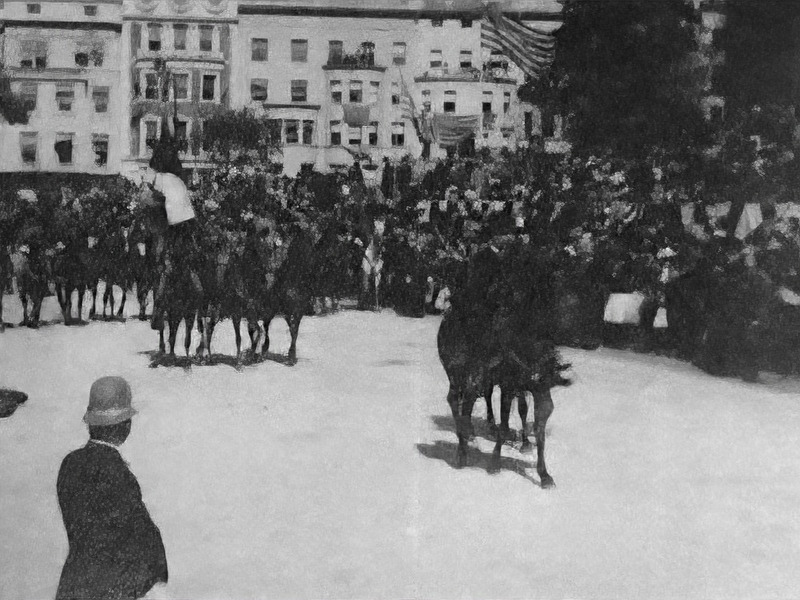} \\
        \end{tabular}%
      }
      \subcaption{Top: original video frames. Bottom: the restored frames.}
      \label{fig:teaser:qual}
    \end{subfigure}\hspace{0.4em}
    \begin{subfigure}[b]{0.35\linewidth}
      \centering
      % Vertically center the plot w.r.t. the 2x3 grid on the left.
      \begin{minipage}[c][3.6cm][c]{\linewidth}
        \centering
        \includegraphics[height=3.6cm,keepaspectratio]{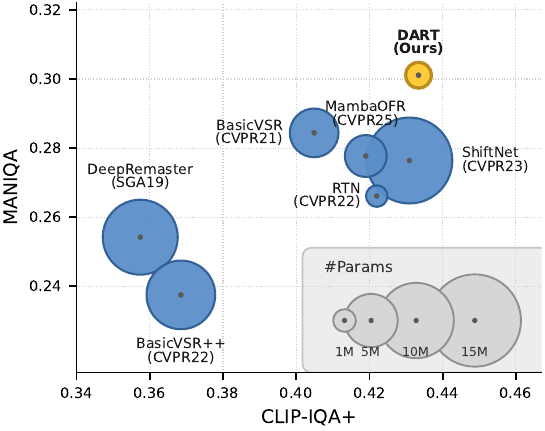}
      \end{minipage}
      \subcaption{Performance gain}
      \label{fig:teaser:bubble}
    \end{subfigure}
    \caption{(a) Real-world old films restored by our method (b) DART beats prior state-of-the-art while remaining efficient.}
    \label{fig:teaser}
  \end{figure}
}%

% ===============================================================
\begin{abstract}
Archival film restoration is a challenging problem because historical footage contains compound degradations such as scratches, dust, blur, noise, flicker, and photometric aging, while clean reference videos are unavailable. Existing video restoration methods largely treat these degradations implicitly, reconstructing frames without explicit knowledge of where damage occurs or how severe it is. We propose DART, a degradation-aware recurrent transformer for archival film restoration. DART predicts and propagates a soft defect mask through time, using it to guide temporal fusion and condition the restoration network on both damage location and severity. This makes the restoration process explicitly aware of film artifacts rather than relying only on reconstruction losses. Experiments on real archival benchmarks show that DART improves no-reference perceptual quality over prior restoration architectures while remaining compact and efficient, producing cleaner and more temporally consistent restorations of structured film damage.
\end{abstract}

% ===============================================================
\section{Introduction}
\label{sec:intro}

Archival film is an irreplaceable visual record of cultural, historical, and artistic memory. 
However, much of this footage has survived only in degraded form. 
Old films commonly contain a mixture of structural defects, such as scratches, dust, cracks, and blotches, together with photometric degradation, blur, flicker, exposure instability, and artifacts introduced during digitization. 
Restoring such material is therefore not a standard denoising or enhancement problem. A model must distinguish genuine scene content from damage, preserve temporal coherence, and operate in a setting where clean reference videos are unavailable for real archival footage.
\footnote{Code: \url{https://github.com/TytanMikJas/DART-Degradation-Aware-Recurrent-Transformer}}

Recent video restoration methods have made substantial progress by exploiting temporal information across frames through alignment, recurrent propagation, and transformer-based reconstruction~\cite{chan2021basicvsr,chan2022basicvsrpp,liang2022rvrt,li2023simple}. 
These ideas have also been adapted to old-film restoration, where models must handle stronger and more diverse artifacts than in standard video enhancement~\cite{deepremaster,wan2022bringing,lin2024restoring,mao2025making}. 
However, most existing restorers still treat degradation only implicitly, reconstructing clean frames without being told where damage occurs or how severe it is, so defect localization is learned only indirectly through the reconstruction loss rather than as a supervised signal.
Recent supervised detection approaches address part of this issue, but are limited to structural damage and are usually separated from the restoration model itself~\cite{liu2026restoration}.
This limits the model's ability to distinguish true film damage from valid scene structures and motivates a restoration framework that is explicitly degradation-aware.

This lack of explicit damage awareness is especially problematic for archival footage.
A scratch, a fence post, a branch, and a thin object boundary can all produce similar high-contrast patterns yet require different restoration behavior.
Conversely, severe localized degradation may require the model to rely more heavily on temporal evidence from neighboring frames. 
A restoration model should therefore reason explicitly about \emph{where} the footage is damaged and \emph{how strongly} to intervene, not just reconstruct plausible images.

We propose \textbf{DART}, a \textbf{D}egradation-\textbf{A}ware \textbf{R}ecurrent \textbf{T}ransformer for archival film restoration.
It predicts a continuous soft defect mask, propagates this mask through time, and uses it to guide recurrent fusion between the current observation and the aligned temporal state.
Crucially, unlike prior restorers that also rely on defect masks but learn them only implicitly through the reconstruction loss, DART supervises this mask directly against ground-truth defect locations, so damage localization is optimized explicitly during training rather than emerging as a by-product of reconstruction.
In addition, DART summarizes the predicted mask and residual evidence into a global degradation condition, allowing the restoration transformer to adapt its behavior to the severity of the current frame.
In summary, our main contributions are:
\begin{itemize}
    \item We introduce DART, a degradation-aware recurrent restoration framework for compound archival film degradation, shifting old-film restoration from passive reconstruction to explicit damage-aware processing.

    \item We propose a multi-scale Dilation Pyramid MaskNet trained with direct continuous-mask supervision, enabling the model to localize both the position and severity of film artifacts.

    \item We condition the restoration backbone on degradation through AdaLN-Zero modulation driven by the predicted mask and residual evidence, allowing the network to adapt its restoration behavior to the severity of each frame.

    \item We achieve state-of-the-art results on real archival-film benchmarks while maintaining a compact model.
\end{itemize}

% ===============================================================
\section{Related Work}
\label{sec:related}
% Video restoration refers to recovering high-quality clips based on a corrupted sequence of frames and is a family of problems such as super resolution \cite{liang2021swinir,chan2021basicvsr,chan2022basicvsrpp}, colorization \cite{lei2019fully,zhang2019deep}, denoising \cite{tassano2020fastdvdnet,sheth2021unsupervised}, and archival video restoration \cite{deepremaster,wan2020bringing,mao2025making}. \\
\noindent\textbf{Temporal modeling architectures.} A central challenge in video restoration is generating details while maintaining consistency across frames. Early approaches incorporated temporal information by aggregating neighboring frames. FastDVDnet \cite{tassano2020fastdvdnet} and UDVD \cite{sheth2021unsupervised} concatenate adjacent frames to enhance the target frame, while Lei et al. \cite{lei2019fully} introduces temporal consistency constraints to reduce frame-to-frame variations. Later methods modeled frame correspondences through alignment and propagation mechanisms. EDVR \cite{wang2019edvr} and TDAN \cite{tian2020tdan} introduced deformable alignment modules to handle complex motion without explicit optical flow \cite{horn1981determining}. Recurrent architectures improved temporal information exchange by propagating features across the sequence. BasicVSR and BasicVSR++ \cite{chan2021basicvsr,chan2022basicvsrpp} demonstrated bidirectional feature propagation, while later works explored propagation in latent representations \cite{wan2020bringing,lin2024restoring}. RVRT \cite{liang2022rvrt} combined recurrent propagation with guided deformable attention to capture long-range dependencies and spatial correspondences. Closer to our setting, RTN~\cite{wan2022bringing} brought the recurrent-transformer paradigm to old-film restoration, propagating a hidden state to carry clean content across damaged frames. To reduce the latency of recurrent approaches, recent methods investigate more efficient temporal fusion. DeepRemaster \cite{deepremaster} introduced reference-based attention in the bottleneck of an autoencoder, enabling joint processing of multiple input frames. More recently, Mao et al. \cite{mao2025making} adopted Mamba-based state-space modeling \cite{gu2023mamba} as an efficient alternative to self-attention, applied along the spatial dimension. Despite their strong temporal modeling, these methods remain \emph{degradation-unaware}: they reconstruct clean frames without ever being told where a frame is damaged or how severely, and none conditions its restoration on the degradation itself. Making restoration explicitly damage-aware first requires localizing the damage. \\

\noindent\textbf{Degradation segmentation.} Explicit mask prediction is a critical strategy in restoration used to isolate corrupted regions and prevent the blurring of clean textures. Historically, masking has been very common in dehazing \cite{ngo2021haziness,liu2025hdsa} to estimate spatial density maps that point out exactly where haze is concentrated. Similarly, it is intuitively used to segment out distinct artifacts like rain streaks \cite{guan2025clip}, or to mark extremely dark, information-sparse spots in low-light image enhancement \cite{hou2023global}. While these spatial priors are effective for static scenes, masking becomes similarly useful for old video enhancement \cite{lin2024restoring,mao2025making}. 
Historical films suffer from unstructured, temporally random defects like scratches and dust. To maintain temporal consistency and preserve structural fidelity, modern architectures use degradation masks to isolate these artifacts. Methods such as \cite{wan2022bringing, lin2024restoring,mao2025making} rely on feature map differences to estimate these masks. However, because natural scene dynamics also alter these features, the network often misidentifies camera or object motion as true film degradations. A recent detector~\cite{liu2026restoration} instead supervises damage segmentation directly, avoiding this confusion, but it targets only binary, structural defects and runs as a separate detection stage decoupled from the restoration network. 
DART instead supervises a \emph{continuous} defect mask end-to-end within the restoration network, capturing not only where damage occurs but how severe it is, across the full range of compound film degradation rather than binary structural damage alone. By anchoring mask generation to ground-truth defects, our approach targets only actual artifacts, disentangling them from scene motion.

% ===============================================================
\section{Method}
\label{sec:method}

\subsection{Overview}
\label{sec:method:overview}
\begin{figure}[tbp]
  \centering
  \includegraphics[width=\linewidth]{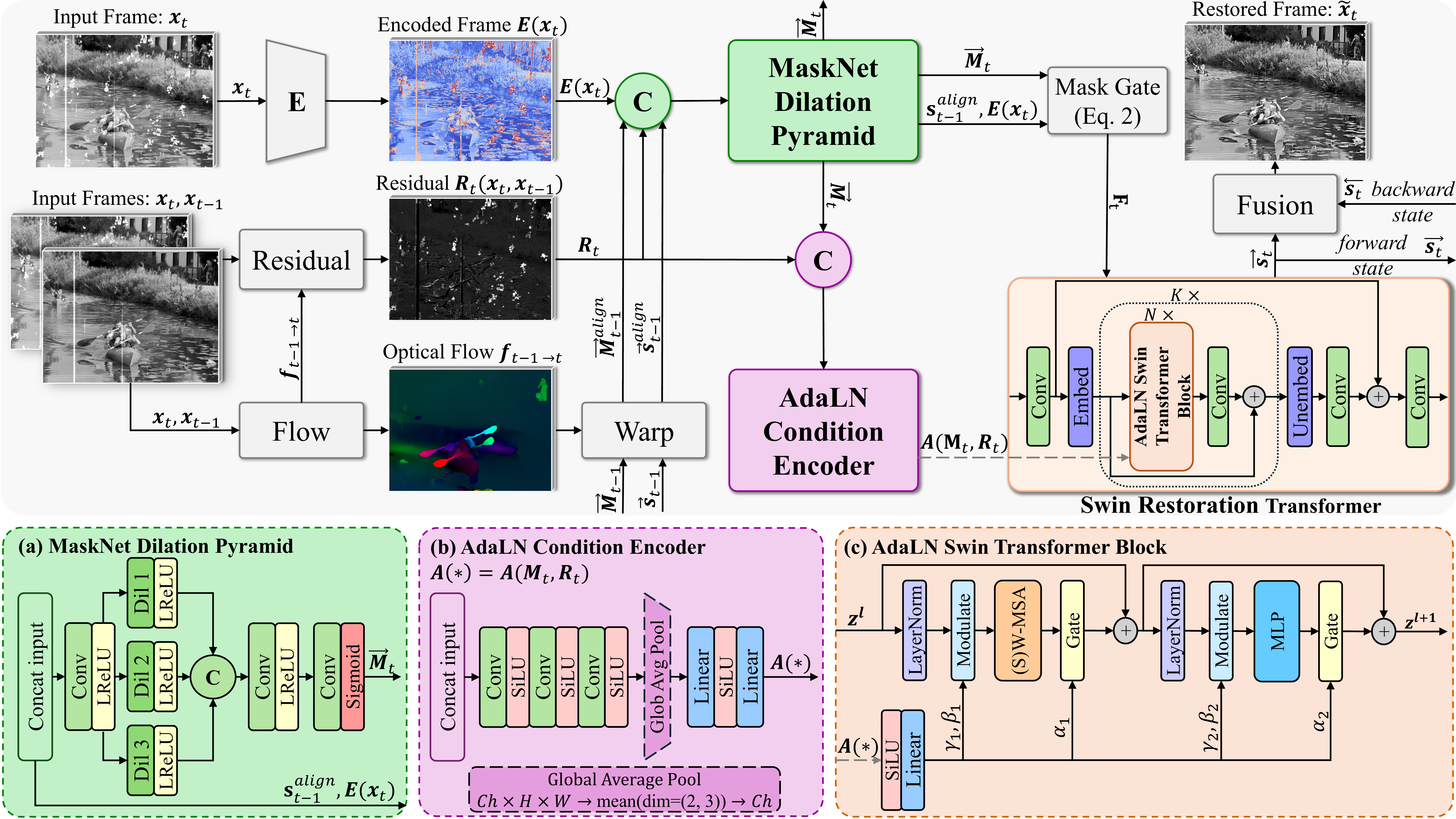}
  \caption{Overview of the proposed DART architecture in one recurrent forward step. C denotes the concatenation operation. The upper portion of the figure illustrates the general flow of the framework, feature alignment and fusion.}
  \label{fig:arch}
\end{figure}

DART restores a $T$-frame clip $\{\mathbf{x}_1,\dots,\mathbf{x}_T\}$ with a bidirectional recurrent network. Each clip is processed once forward, $\mathbf{x}_1\!\rightarrow\!\mathbf{x}_T$, and once backward, $\mathbf{x}_T\!\rightarrow\!\mathbf{x}_1$; the resulting hidden states $\overrightarrow{\mathbf{s}}_t$ and $\overleftarrow{\mathbf{s}}_t$ are merged by a final Fusion convolution into the restored frame $\tilde{\mathbf{x}}_t$. \Cref{fig:arch} depicts a single forward step; the backward pass repeats the same computation in reverse temporal order.

A shared encoder $E$ maps the current frame to a feature $E(\mathbf{x}_t)$, and a frozen RAFT~\cite{teed2020raft} module estimates the optical flow $f_{t-1\to t}$ between $\mathbf{x}_{t-1}$ and $\mathbf{x}_t$. A Warp operator $W(\cdot,\cdot)$ resamples its first argument along a given flow field; applied with $f_{t-1\to t}$ to the previous hidden state $\mathbf{s}_{t-1}$, the previous soft defect mask $\mathbf{M}_{t-1}$, and the previous frame $\mathbf{x}_{t-1}$, it brings all three into alignment with the current coordinate system, yielding $\mathbf{s}^{\text{align}}_{t-1}$, $\mathbf{M}^{\text{align}}_{t-1}$, and a warped previous frame. Comparing the warped previous frame against $\mathbf{x}_t$ gives a residual indicator $\mathbf{R}_t$ that estimates where the frame has changed or is occluded,
\begin{equation}
  \mathbf{R}_t = \big| W(\mathbf{x}_{t-1}, f_{t-1 \to t}) - \mathbf{x}_t \big| .
  \label{eq:residual}
\end{equation}
The soft mask $\mathbf{M}_t$, produced by the MaskNet introduced next, then gates a temporal fusion of the encoded current frame with the aligned hidden state into the representation $\mathbf{F}_t$ that is passed on to the restoration backbone,
\begin{equation}
  \mathbf{F}_t = E(\mathbf{x}_t)\odot(1-\mathbf{M}_t) + \mathbf{s}^{\text{align}}_{t-1}\odot\mathbf{M}_t ;
  \label{eq:fusion}
\end{equation}
wherever $\mathbf{M}_t$ is close to $1$, the network prefers the historical, presumably undamaged state over the current, possibly corrupted observation.

% --- Fig 2: DART architecture (reused, the one good asset) ---
\subsection{Dilation Pyramid MaskNet}
\label{sec:method:masknet}
MaskNet is the component that decides, at every timestep, which regions of the current frame should be trusted and which should instead be filled in from the aligned hidden state $\mathbf{s}^{\text{align}}_{t-1}$. A shallow, directly-connected predictor, the design used by RTN~\cite{wan2022bringing} and left architecturally unchanged in MambaOFR~\cite{mao2025making}, has a receptive field too small to tell genuine scratches apart from thin scene structure such as fence posts, branches, or drifting smoke: both produce a similarly narrow, high-contrast response at the pixel level, so a predictor that only ever looks at a small local window cannot distinguish them.

DART replaces this shallow predictor with a multi-scale \emph{MaskNet Dilation Pyramid} (bottom of \cref{fig:arch} (a)). A shared convolutional stem branches into three parallel paths with dilation rates $d\in\{1,2,3\}$; their outputs are concatenated and merged by a fused head. Because the three branches observe the same input at three different effective receptive fields, the network can confirm a candidate defect against a wider spatial context before committing to it, rather than reacting to a single narrow window in isolation. \Cref{sec:method:propagation} gives the pyramid's full input and the resulting mask $\mathbf{M}_t$; combined with the direct supervision introduced in \cref{sec:method:loss}, this wider receptive field is what lets DART's mask track genuine film damage rather than scene geometry, as \cref{sec:exp:mask} shows on real archival footage.

\subsection{Recurrent Soft-Mask Propagation}
\label{sec:method:propagation}
Predicting $\mathbf{M}_t$ from scratch at every timestep, as RTN does, means that a persistent defect such as a vertical scratch must be independently re-discovered in every single frame, a form of temporal amnesia that wastes model capacity and produces flickering, temporally-inconsistent mask predictions. Following the propagation mechanism introduced in MambaOFR~\cite{mao2025making}, DART instead flow-warps the previous soft mask into the current coordinate frame and feeds it back into the Dilation Pyramid alongside the encoded current frame, the aligned hidden state, and the residual indicator:
\begin{equation}
  \mathbf{M}_t = \mathcal{M}\!\left(E(\mathbf{x}_t)\,\Vert\,W(\mathbf{s}_{t-1},f_{t-1\to t})\,\Vert\,\mathbf{R}_t\,\Vert\,W(\mathbf{M}_{t-1},f_{t-1\to t})\right),
  \label{eq:mask}
\end{equation}
where $\Vert$ denotes channel-wise concatenation and $\mathcal{M}$ the Dilation Pyramid of \cref{sec:method:masknet}. This temporal feedback allows DART to track persistent defects across the clip. As a result, the mask supervision (\cref{sec:method:loss}) acts on a temporally coherent sequence rather than relying on independent per-frame estimates.

\subsection{AdaLN-Zero Degradation Conditioning}
\label{sec:method:adaln}
The mask-gated fusion of \cref{eq:fusion} operates spatially: pixel by pixel, it decides whether to trust the current frame or the historical state. Nothing in that operation tells the restoration backbone \emph{how severely}, overall, the current frame is damaged, so a lightly speckled frame and a heavily cracked one would otherwise be processed by identical transformer weights. DART closes this gap with global degradation conditioning.

A Condition Encoder first concatenates the newly predicted mask $\mathbf{M}_t$ with the residual indicator $\mathbf{R}_t$ and passes the result through three alternating Conv/SiLU layers, a Global Average Pool, and a two-layer MLP, collapsing the spatial degradation evidence into a single condition vector
\begin{equation}
  A(*) = A(\mathbf{M}_t, \mathbf{R}_t).
  \label{eq:condition}
\end{equation}
This vector then modulates every block of the Swin restoration backbone. Each \emph{AdaLN Swin Transformer Block} maps an input feature $\mathbf{z}^l$ to $\mathbf{z}^{l+1}$ by projecting $A(*)$, through a shared SiLU-Linear layer, into per-block modulation parameters $(\gamma_1,\beta_1,\gamma_2,\beta_2)$ and gate scalars $(\alpha_1,\alpha_2)$. A normalized feature $\mathbf{x}$ is scaled and shifted by
\begin{equation}
  \mathrm{Modulate}(\mathbf{x},\gamma,\beta) = \mathbf{x}\odot(1+\gamma) + \beta,
  \label{eq:modulate}
\end{equation}
and the block's two branches, windowed self-attention and an MLP, are each modulated, applied, and re-scaled by their own gate before being added back to the residual stream:
\begin{equation}
\begin{aligned}
    \mathbf{z}^{l}_{in} &= \mathbf{z}^l + \alpha_1 \odot \text{(S)W-MSA}\big(\text{Modulate}(\text{LayerNorm}(\mathbf{z}^l), \gamma_1, \beta_1)\big), \\
    \mathbf{z}^{l+1} &= \mathbf{z}^{l}_{in} + \alpha_2 \odot \text{MLP}\big(\text{Modulate}(\text{LayerNorm}(\mathbf{z}^{l}_{in}), \gamma_2, \beta_2)\big).
\end{aligned}
  \label{eq:adaln-block}
\end{equation}
Here $\text{(S)W-MSA}$ denotes (shifted) window multi-head self-attention, the standard Swin building block~\cite{liang2021swinir} whose window partitioning shifts between consecutive blocks so that information is exchanged across window boundaries, and $\text{MLP}$ a two-layer feed-forward network.
Following the AdaLN-Zero initialization strategy of~\cite{Peebles2022DiT}, the final linear layer that produces $(\gamma,\beta,\alpha)$ is initialized to zero, so every block begins training as an identity mapping and only gradually learns to modulate its features as $A(*)$ becomes informative, which we find stabilizes early training considerably. In effect, the same restoration weights are reused across the whole clip, but the network is free to focus its attention in proportion to how damaged the current frame actually is. This mechanism makes DART's restoration core, and not merely its mask, degradation-aware.

\subsection{Training Objective and Mask Supervision}
\label{sec:method:loss}
DART is optimized end-to-end with a combined objective that balances faithful reconstruction against direct supervision of the defect mask.

\noindent\textbf{Reconstruction and adversarial terms.}
A pixel-wise $\mathcal{L}_1$ term penalizes deviation from the clean ground-truth frame $\mathbf{y}_t$, averaged over the temporal window, and a VGG19-based perceptual term $\mathcal{L}_{perc}$~\cite{johnson2016perceptual} sharpens texture; together they form $\mathcal{L}_{rec} = \mathcal{L}_1 + \lambda_{perc}\mathcal{L}_{perc}$. A Temporal-PatchGAN discriminator~\cite{8100115} built from 3D convolutions is trained with a hinge loss to distinguish restored clips from clean ones, and DART is trained to fool it via the adversarial term $\mathcal{L}_{adv}$.

\noindent\textbf{Mask supervision.}
Reconstruction alone gives MaskNet only an indirect, weak training signal, since a mask can be substantially wrong while the reconstructed frame still looks acceptable. DART instead supervises the mask directly against a continuous ground-truth mask $\mathbf{M}^{gt}$, exported alongside the synthetic training pairs (\cref{sec:exp:setup}), applied to the raw pre-sigmoid MaskNet logits $\mathbf{Z}$ in both the forward and backward propagation passes; head frames, for which no temporal gate yet exists, are excluded. Because defect pixels are sparse, the binary cross-entropy term is class-balanced with a positive weight $w_{pos} = \mathrm{clamp}(N_{neg}/N_{pos},\, w_{min},\, w_{max})$, where $N_{pos}$ and $N_{neg}$ count positive and negative ground-truth pixels:
\begin{equation}
    \mathcal{L}_{BCE}^{(p)} = \mathrm{BCEWithLogits}\!\left(\mathbf{Z}^{(p)},\, \mathbf{M}^{gt,(p)},\, w_{pos}\right).
    \label{eq:bce}
\end{equation}
A differentiable Dice term further rewards structural overlap between the predicted soft mask $\mathbf{M}=\sigma(\mathbf{Z}^{(p)})$ and the ground truth:
\begin{equation}
    \mathcal{L}_{Dice}^{(p)} = 1 - \frac{2\sum(\mathbf{M}\odot\mathbf{M}^{gt,(p)}) + \epsilon}{\sum\mathbf{M} + \sum\mathbf{M}^{gt,(p)} + \epsilon}.
    \label{eq:dice}
\end{equation}
The two terms are combined and averaged over both propagation directions $p\in\{\leftarrow,\rightarrow\}$:
\begin{equation}
    \mathcal{L}_{mask} = \frac{1}{2}\sum_{p\in\{\leftarrow,\rightarrow\}}\left(\mathcal{L}_{BCE}^{(p)} + \lambda_{dice}\,\mathcal{L}_{Dice}^{(p)}\right).
    \label{eq:mask-loss}
\end{equation}
Unlike prior methods that infer defect locations indirectly through reconstruction, this explicit formulation forces MaskNet to learn the exact position and severity of degradations directly from the ground truth.

\noindent\textbf{Full objective.}
The three terms are combined as
\begin{equation}
  \mathcal{L}_{total} = \lambda_{rec}\,\mathcal{L}_{rec}
  + \lambda_{adv}\,\mathcal{L}_{adv}
  + \lambda_{mask}\,\mathcal{L}_{mask},
  \label{eq:total}
\end{equation}
with the loss weights $\lambda_{rec}, \lambda_{adv}, \lambda_{mask}, \lambda_{dice}$ reported alongside the remaining training settings in \Cref{tab:impl}.

% ===============================================================
\section{Experiments}
\label{sec:exp}

\begin{table}[tbp]
  \centering
  \footnotesize
  \caption{Training settings and loss weights of DART.}
  \label{tab:impl}
  \setlength{\tabcolsep}{6pt}
  \renewcommand{\arraystretch}{1.05}
  \begin{tabular}{@{}ll@{}}
    \toprule
    Setting & Value \\
    \midrule
    Optimizer & Adam ($\beta_1{=}0.9,\beta_2{=}0.99$) \\
    Learning rate & $2\times10^{-4}$ (linear decay) \\
    Epochs & 20 \\
    Window / crop / batch & 7 frames / $256{\times}256$ / 1 clip \\
    Precision & FP16 (mixed) \\
    $\lambda_{rec},\lambda_{adv},\lambda_{mask},\lambda_{dice}$ & $1.0,\ 0.01,\ 0.5,\ 0.5$ \\
    \bottomrule
  \end{tabular}
\end{table}

\subsection{Datasets, Metrics, and Implementation}
\label{sec:exp:setup}
\noindent\textbf{Training data.}
DART is trained on REDS~\cite{Nah_2019_CVPR_Workshops_REDS} \texttt{train\_sharp} (720p, 24\,fps). During training, a synthetic degradation pipeline, adapted from AbsoluteDegradation~\cite{absolutedegradation_eval_2026}, generates training pairs online, yielding for each $256\times256$ crop a ground-truth sharp frame, a synthetically degraded version of it, and a continuous artifact mask: the mask outlines exactly where the synthetic defects were applied and undergoes the same degradation as the degraded frame itself. This mask provides the supervision target for $\mathcal{L}_{mask}$ (\cref{eq:mask-loss}). The official 30 REDS \texttt{val\_sharp} clips are reserved as a synthetic, full-reference proxy test set, used for the ablation study in \cref{sec:exp:ablation} and for the full-reference comparison in the supplementary material.

\noindent\textbf{Real-world evaluation.}
No clean references exist for genuine archival footage, so evaluation on real data is necessarily no-reference. AbsoluteDegradation~\cite{absolutedegradation_eval_2026}, a 13{,}252-frame, lossless-PNG corpus of public-domain archival footage (1896--1918), is the primary target domain; SRWOV~\cite{mao2025making}, 216 lower-resolution, JPEG-compressed real-world clips, serves as a secondary benchmark.

\noindent\textbf{Metrics.}
Since real archival footage has no clean reference, restoration fidelity cannot be measured on it directly. We therefore score real footage with three no-reference metrics learned from human quality ratings: CLIPIQA+~\cite{wang2023exploring}, MUSIQ~\cite{ke2021musiq}, and MANIQA~\cite{yang2022maniqa}. Each rates perceptual quality and naturalness directly, the only judgment available without a clean target. Their different architectures (a CLIP vision--language prior, a multi-scale transformer, and multi-dimension attention) provide complementary perceptual assessments and make agreement among them unlikely to be an artifact of any single model. Full-reference PSNR, SSIM~\cite{wang2004image}, and LPIPS~\cite{zhang2018unreasonable} can only be computed on the synthetic proxy set, whose degradations differ in distribution from real film, so a high proxy score does not guarantee good results on real footage (\cref{sec:exp:ablation}).

\noindent\textbf{Implementation.}
DART is implemented in PyTorch and trained on two NVIDIA A100 GPUs with FP16 mixed precision, using a frozen RAFT~\cite{teed2020raft} flow estimator. \Cref{tab:impl} lists the remaining training settings and loss weights. Every competing architecture in \cref{sec:exp:sota} is retrained from scratch on identical data using its own published hyperparameters, so all comparisons in this section share a common training source and evaluation protocol.

\subsection{Comparison with State of the Art}
\label{sec:exp:sota}
\noindent\textbf{Quantitative results.}
\Cref{tab:model_comparison} compares DART against BasicVSR~\cite{chan2021basicvsr}, BasicVSR++~\cite{chan2022basicvsrpp}, ShiftNet~\cite{li2023simple}, DeepRemaster~\cite{deepremaster}, RTN~\cite{wan2022bringing}, and MambaOFR~\cite{mao2025making}, all retrained on identical data. DART outperforms all baselines on every metric on AbsoluteDegradation. On SRWOV it remains the best on MUSIQ and MANIQA, the two metrics that correlate most closely with human perceptual judgment, and is edged out only on CLIPIQA+, where ShiftNet scores narrowly higher (0.458 vs.\ 0.449). DART outperforms RTN, the architecture whose Swin restoration backbone it shares, on every metric across both datasets; therefore, this consistent margin isolates what the degradation-aware additions themselves contribute, rather than reflecting a larger or differently-trained network.

\begin{table}[t]
    \centering
    \footnotesize
    \caption{Quantitative comparison of restoration architectures, all trained on the AbsoluteDegradation pipeline and evaluated on the \textbf{AbsoluteDegradation} and \textbf{SRWOV}~\cite{mao2025making} datasets. Bold marks the best score, and underline marks the second best, in each metric column. The input frames row denotes the unrestored degraded footage and serves as a reference floor.}
    \label{tab:model_comparison}
    \setlength{\tabcolsep}{0pt}
    \renewcommand{\arraystretch}{0.92}
    \begin{tabularx}{\linewidth}{@{}
        >{\raggedright\arraybackslash}p{0.205\linewidth}|
        >{\centering\arraybackslash\hsize=1.17\hsize}X
        >{\centering\arraybackslash\hsize=0.82\hsize}X
        >{\centering\arraybackslash\hsize=1.01\hsize}X|
        >{\centering\arraybackslash\hsize=1.17\hsize}X
        >{\centering\arraybackslash\hsize=0.82\hsize}X
        >{\centering\arraybackslash\hsize=1.01\hsize}X@{}}
        \toprule
        & \multicolumn{3}{c}{\textbf{AbsoluteDegradation Dataset}} & \multicolumn{3}{c}{\textbf{SRWOV Dataset}} \\
        \cmidrule(lr){2-4}\cmidrule(lr){5-7}
        \textbf{Model} & CLIPIQA+~$\uparrow$ & MUSIQ~$\uparrow$ & MANIQA~$\uparrow$ &
        CLIPIQA+~$\uparrow$ & MUSIQ~$\uparrow$ & MANIQA~$\uparrow$ \\
        \midrule
        Input frames & 0.2483 & 26.8418 & 0.1619 & 0.3890 & 42.8591 & 0.2307 \\
        \midrule
        \mbox{BasicVSR\cite{chan2021basicvsr}}           & 0.4049 & \underline{49.5797} & \underline{0.2844} & 0.4086 & 57.6442 & 0.3104 \\
        \mbox{BasicVSR++\cite{chan2022basicvsrpp}}       & 0.3685 & 42.9144 & 0.2375 & 0.4428 & 58.0442 & 0.3231 \\
        \mbox{ShiftNet\,\cite{li2023simple}}               & \underline{0.4309} & 46.7724 & 0.2764 & \textbf{0.4578} & 57.9334 & \underline{0.3336} \\
        \mbox{DeepRemaster\cite{deepremaster}}           & 0.3574 & 44.7540 & 0.2542 & 0.3722 & 53.5473 & 0.2637 \\
        \mbox{RTN\cite{wan2022bringing}}                 & 0.4220 & 46.8737 & 0.2661 & 0.4441 & \underline{58.4585} & 0.3321 \\
        \mbox{MambaOFR\cite{mao2025making}}              & 0.4190 & 47.2436 & 0.2777 & 0.4402 & 57.3551 & 0.3210 \\
        DART (Ours)                                & \textbf{0.4334} & \textbf{51.2304} & \textbf{0.3011} & \underline{0.4488} & \textbf{59.4263} & \textbf{0.3427} \\
        \bottomrule
    \end{tabularx}
\end{table}

\noindent\textbf{Qualitative results.}
\Cref{fig:qual} tracks three consecutive frames from an archival newsreel clip. BasicVSR++ and MambaOFR leave the emulsion out of focus and let speckle persist across all three frames; RTN removes some noise but leaves faint vertical scratch tracks; DART alone produces a stable, sharp, and temporally consistent reconstruction. \Cref{fig:model_comparison_sea_trip_market_race} extends the comparison to three further clips: across all of them, DART suppresses structured defects, scratches, and dust while preserving genuine scene texture more faithfully than the generic video-restoration baselines, which either under-react to structured damage or over-sharpen it into the scene.

\noindent\textbf{Efficiency.}
DART is not the fastest model in \cref{tab:eff}: DeepRemaster~\cite{deepremaster} runs about $18\times$ faster, but trails DART by $6.5$ MUSIQ. DART is the second-smallest model, at $6.6$M parameters (only $1.3$M trainable beyond the frozen RAFT), among the most memory-efficient at $0.35$\,GB, and runs at a speed comparable to BasicVSR, BasicVSR++, and RTN. It dominates MambaOFR~\cite{mao2025making} on every axis and improves on ShiftNet~\cite{li2023simple} in parameters, memory, and speed. DART thus trades peak throughput for the highest quality at a compact size, which suits offline archival restoration.

\begin{table}[t]
  \centering
  \footnotesize
  \caption{Computational cost and restoration quality, measured per frame on a
  $1{\times}7{\times}3{\times}256{\times}256$ clip (\textbf{NVIDIA A100 GPU}, fp32, 30 timed
  iterations after 5 warmup). MUSIQ is from \cref{tab:model_comparison}. Params include the
  shared, frozen RAFT ($5.3$M); DART adds $1.3$M trainable.
  \textbf{Bold}/\underline{underline}: best/second-best.}
  \label{tab:eff}
  \setlength{\tabcolsep}{6pt}
  \renewcommand{\arraystretch}{0.92}
  \begin{tabular}{@{}l|ccccc@{}}
    \toprule
    Method & Params\,(M)\,$\downarrow$ & FLOPs\,(G)\,$\downarrow$ & Mem\,(GB)\,$\downarrow$ & FPS\,$\uparrow$ & MUSIQ\,$\uparrow$ \\
    \midrule
    \mbox{BasicVSR\,\cite{chan2021basicvsr}}     & 9.9  & 910             & \textbf{0.34}  & 27.6           & \underline{49.6} \\
    \mbox{BasicVSR++\,\cite{chan2022basicvsrpp}} & 13.6 & 381 & 0.36           & 30.6           & 42.9 \\
    \mbox{ShiftNet\,\cite{li2023simple}}         & 13.0 & \underline{316} & 1.06           & 21.4           & 46.8 \\
    \mbox{DeepRemaster\,\cite{deepremaster}}     & 9.9  & \textbf{131}    & 0.39           & \textbf{529.3} & 44.8 \\
    \mbox{RTN\,\cite{wan2022bringing}}           & \textbf{6.2} & 349     & \underline{0.35} & \underline{33.8} & 46.9 \\
    \mbox{MambaOFR\,\cite{mao2025making}}        & 8.7  & 442 & 0.64           & 14.5           & 47.2 \\
    DART (Ours)                 & \underline{6.6} & 372  & \underline{0.35}  & 29.5           & \textbf{51.2} \\
    \bottomrule
  \end{tabular}
\end{table}

% --- Fig 4: qualitative comparison on three consecutive Roosevelt frames ---
\begin{figure}[tbp]
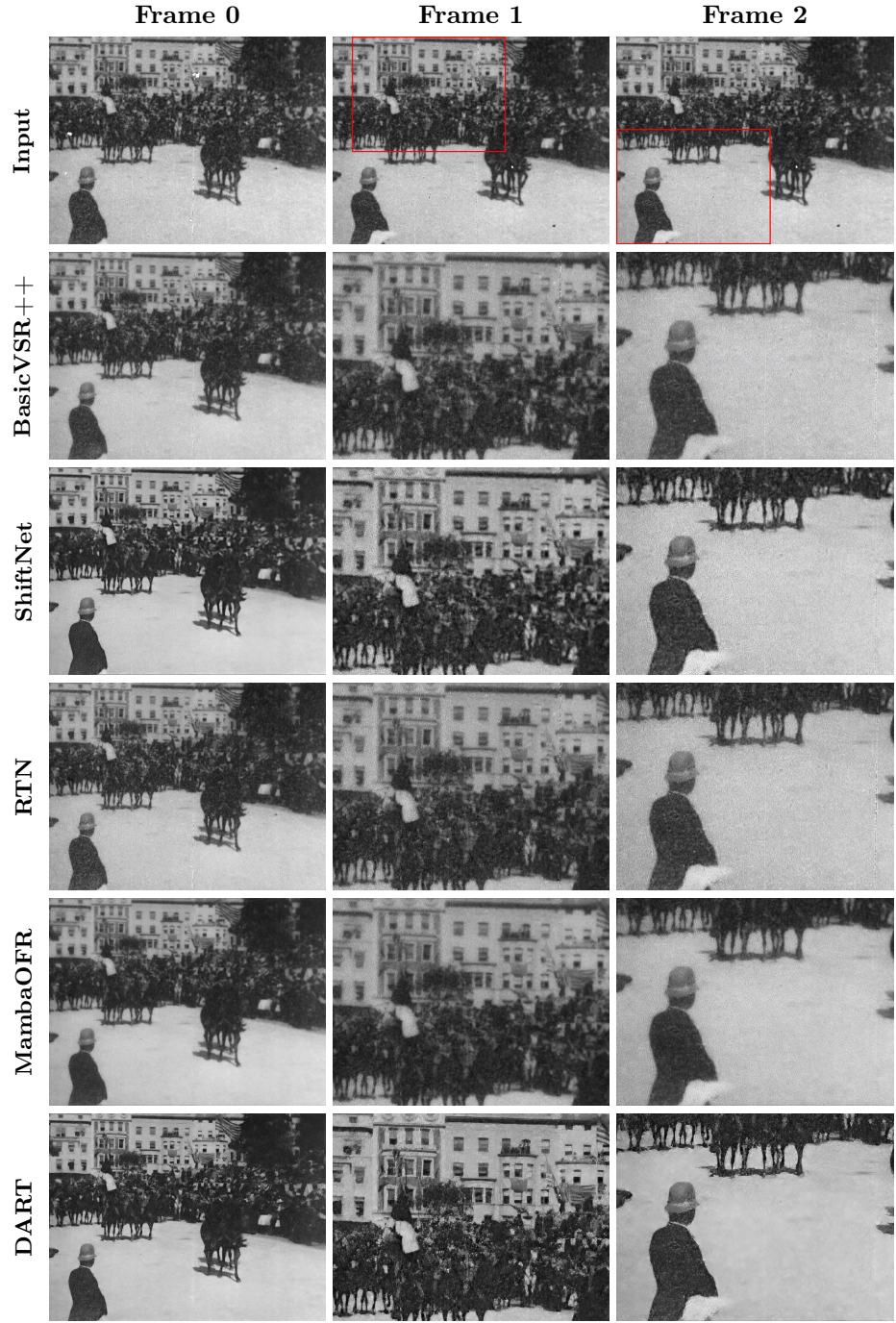

  \centering
  \setlength{\tabcolsep}{0pt}
  \renewcommand{\arraystretch}{1.0}
  \modelcmpgrid{%
    \begin{tabular}{@{}>{\centering\arraybackslash}m{\modelcmplabelwidth}@{\hspace{\modelcmppad}}>{\centering\arraybackslash}m{\modelcmpimgwidth}@{\hspace{\modelcmppad}}>{\centering\arraybackslash}m{\modelcmpimgwidth}@{\hspace{\modelcmppad}}>{\centering\arraybackslash}m{\modelcmpimgwidth}@{}}
      & \normalsize\textbf{Frame 0} & \normalsize\textbf{Frame 1} & \normalsize\textbf{Frame 2} \\[2pt]
      \modelcmptemporalrow{\textbf{Input}}{roosevelt/frame_0/raw_input.jpg}{roosevelt/frame_1/raw_input.jpg}{roosevelt/frame_2/raw_input.jpg}
      \modelcmptemporalrow{\textbf{BasicVSR++}}{roosevelt/frame_0/basicvsrpp.jpg}{roosevelt/frame_1/basicvsrpp.jpg}{roosevelt/frame_2/basicvsrpp.jpg}
      \modelcmptemporalrow{\textbf{ShiftNet}}{roosevelt/frame_0/shiftnet.jpg}{roosevelt/frame_1/shiftnet.jpg}{roosevelt/frame_2/shiftnet.jpg}
      \modelcmptemporalrow{\textbf{RTN}}{roosevelt/frame_0/rtn.jpg}{roosevelt/frame_1/rtn.jpg}{roosevelt/frame_2/rtn.jpg}
      \modelcmptemporalrow{\textbf{MambaOFR}}{roosevelt/frame_0/mambaofr.jpg}{roosevelt/frame_1/mambaofr.jpg}{roosevelt/frame_2/mambaofr.jpg}
      \modelcmptemporalrow{\textbf{DART}}{roosevelt/frame_0/mine_baseline.jpg}{roosevelt/frame_1/mine_baseline.jpg}{roosevelt/frame_2/mine_baseline.jpg}
    \end{tabular}%
  }
  \caption{Qualitative comparison on three consecutive frames of an archival clip. Competing methods leave vertical scratch tracks, over-sharpen emulsion noise, or leave the frame out of focus; DART produces a cleaner, sharper, and more temporally consistent reconstruction across the sequence.}
  \label{fig:qual}
\end{figure}

% --- Fig: qualitative comparison across Sea, Baltimore, Race clips ---
\begin{figure}[tbp]
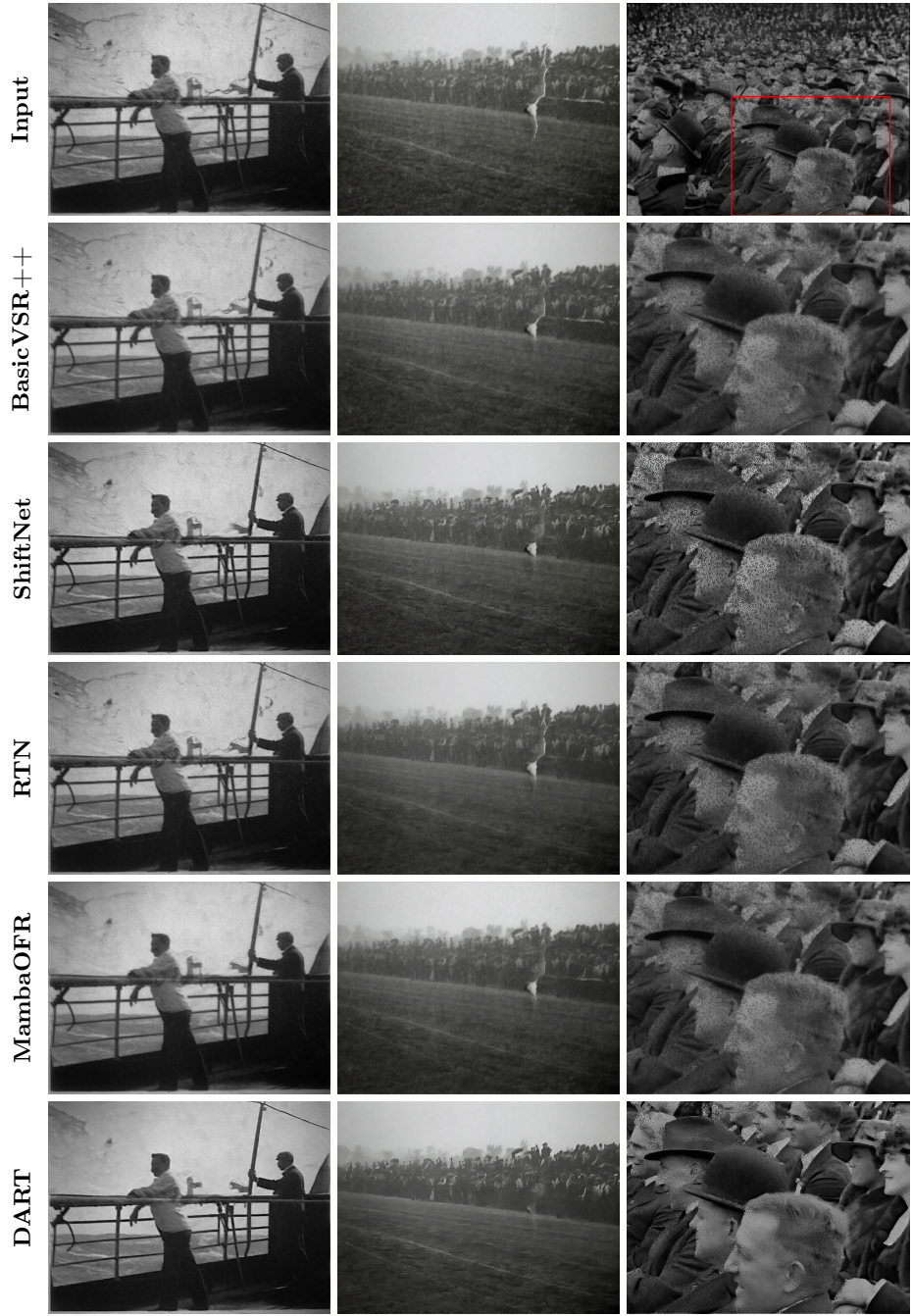

  \centering
  \setlength{\tabcolsep}{0pt}
  \renewcommand{\arraystretch}{1.0}
  {\setlength{\modelcmpgridheight}{0.60\textheight}%
  \modelcmpgrid{%
    \begin{tabular}{@{}>{\centering\arraybackslash}m{\modelcmplabelwidth}@{\hspace{\modelcmppad}}>{\centering\arraybackslash}m{\modelcmpimgwidth}@{\hspace{\modelcmppad}}>{\centering\arraybackslash}m{\modelcmpimgwidth}@{\hspace{\modelcmppad}}>{\centering\arraybackslash}m{\modelcmpimgwidth}@{}}
      \modelcmpthreerow{\textbf{Input}}{sea/raw_input.jpg}{race/raw_input.jpg}{baltimore/frame_2/raw_input.jpg}
      \modelcmpthreerow{\textbf{BasicVSR++}}{sea/basicvsrpp.jpg}{race/basicvsrpp.jpg}{baltimore/frame_2/basicvsrpp.jpg}
      \modelcmpthreerow{\textbf{ShiftNet}}{sea/shiftnet.jpg}{race/shiftnet.jpg}{baltimore/frame_2/shiftnet.jpg}
      \modelcmpthreerow{\textbf{RTN}}{sea/rtn.jpg}{race/rtn.jpg}{baltimore/frame_2/rtn.jpg}
      \modelcmpthreerow{\textbf{MambaOFR}}{sea/mambaofr.jpg}{race/mambaofr.jpg}{baltimore/frame_2/mambaofr.jpg}
      \modelcmpthreerow{\textbf{DART}}{sea/mine_baseline.jpg}{race/mine_baseline_dcn.jpg}{baltimore/frame_2/mine_baseline.jpg}
    \end{tabular}%
  }}
  \caption{Qualitative comparison on archival clips. DART suppresses structured defects and scratches while preserving scene texture more faithfully than the generic video-restoration baselines, which either under-react to structured damage or over-sharpen it into the scene.}
  \label{fig:model_comparison_sea_trip_market_race}
\end{figure}

\subsection{Mask Behavior on Real Footage}
\label{sec:exp:mask}
\begin{figure}[t]
  \centering
  \setlength{\tabcolsep}{0pt}%
  \renewcommand{\arraystretch}{0}%
  \begin{tabular}{@{}cccc@{}}
    {\footnotesize Input} & {\footnotesize RTN} & {\footnotesize MambaOFR} & {\footnotesize DART (Ours)} \\
    \includegraphics[width=0.25\linewidth]{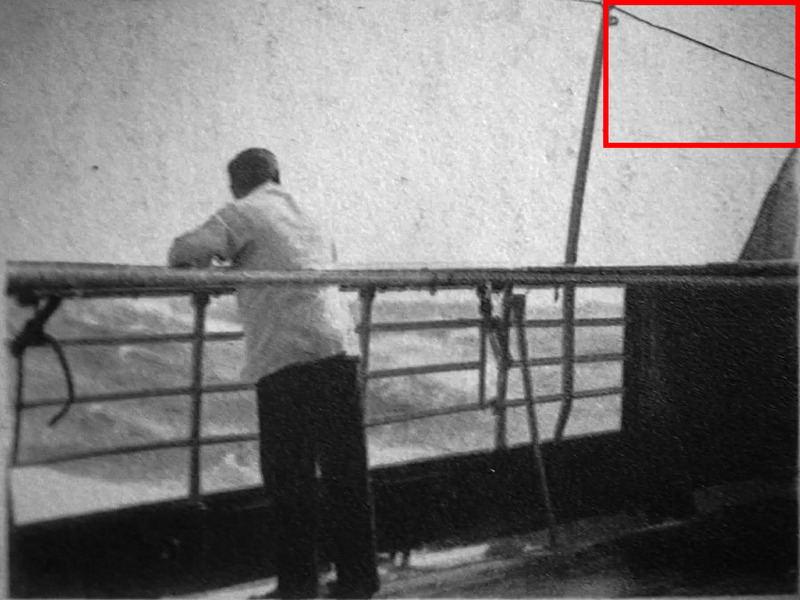} &
    \includegraphics[width=0.25\linewidth]{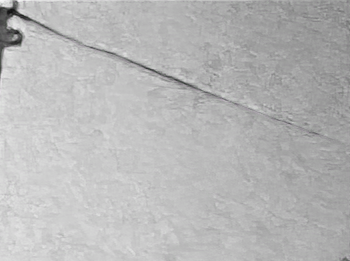} &
    \includegraphics[width=0.25\linewidth]{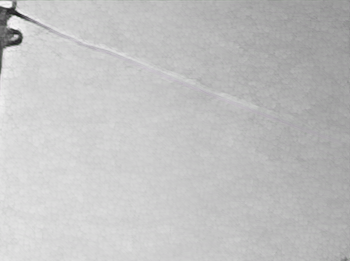} &
    \includegraphics[width=0.25\linewidth]{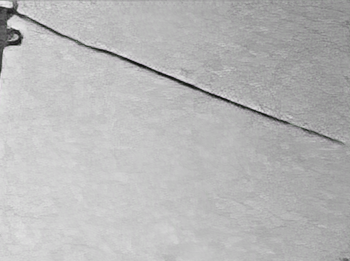} \\
  \end{tabular}
  \caption{RTN and MambaOFR mistake a ship's deck railing, genuine scene structure with a scratch-like profile, for film damage and partially erase it during temporal fusion; DART's mask recognizes the structure and restores it intact.}
  \label{fig:false_positive}
\end{figure}

\begin{figure}[t]
  \centering
  \setlength{\tabcolsep}{0pt}%
  \renewcommand{\arraystretch}{0}%
  \begin{tabular}{@{}cccc@{}}
    {\footnotesize Input} & {\footnotesize RTN} & {\footnotesize MambaOFR} & {\footnotesize DART (Ours)} \\
    \includegraphics[width=0.25\linewidth]{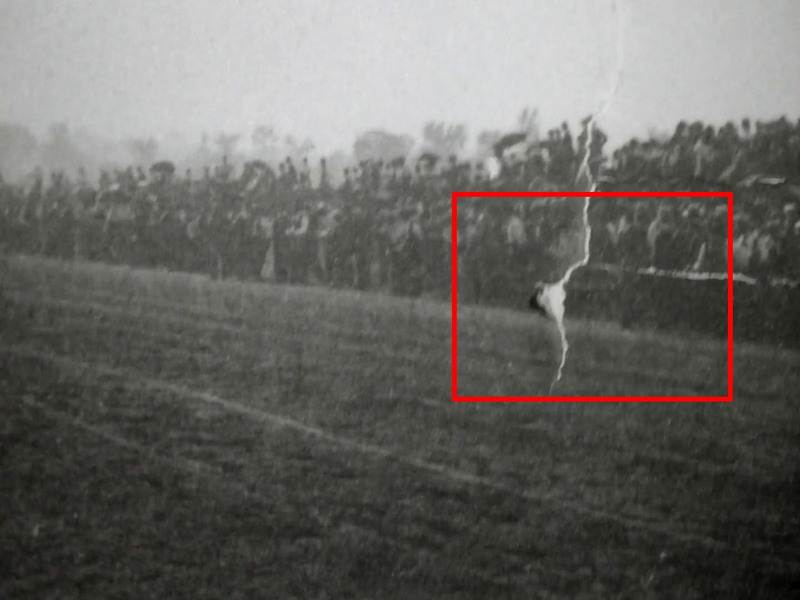} &
    \includegraphics[width=0.25\linewidth]{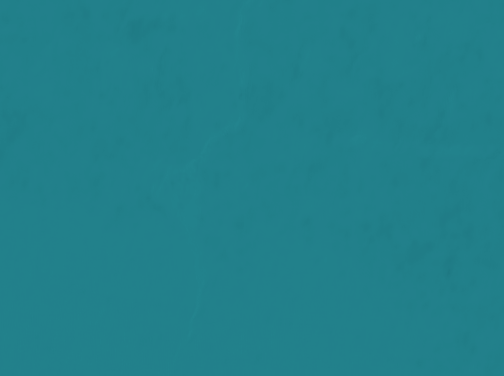} &
    \includegraphics[width=0.25\linewidth]{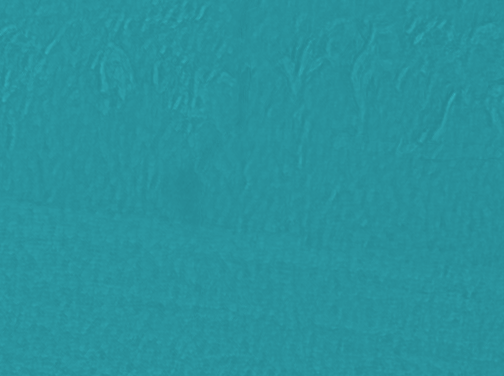} &
    \includegraphics[width=0.25\linewidth]{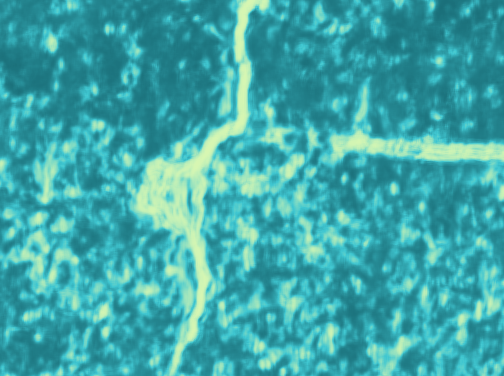} \\
    \includegraphics[width=0.25\linewidth]{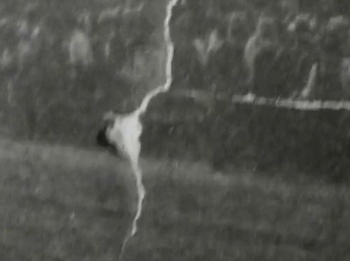} &
    \includegraphics[width=0.25\linewidth]{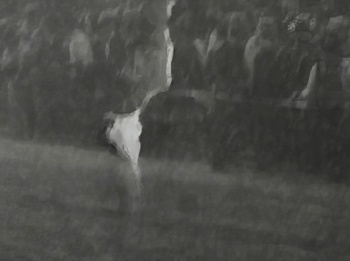} &
    \includegraphics[width=0.25\linewidth]{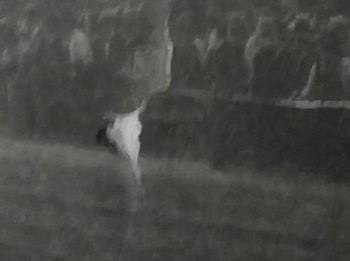} &
    \includegraphics[width=0.25\linewidth]{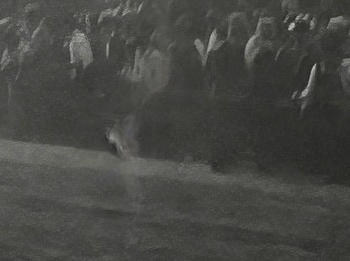} \\
  \end{tabular}
  \caption{Mask activations (top) and restored crops (bottom) for a severe chemical crack in the emulsion. RTN and MambaOFR's unsupervised masks stay essentially flat and fail to localize the defect at all, so it survives largely unrestored; DART's supervised mask activates precisely on the crack, and the corresponding output attenuates it substantially.}
  \label{fig:false_positive_race}
\end{figure}

Because the defect masks in RTN and MambaOFR are never directly supervised, their accuracy on real archival footage, where no ground-truth mask exists to check them against, cannot be inferred from training loss alone. \Cref{fig:false_positive} and \Cref{fig:false_positive_race} illustrate two recurring failure modes of this unsupervised design, both absent from DART.

\Cref{fig:false_positive} isolates a diagonal deck railing on a ship, genuine scene structure with exactly the narrow, high-contrast profile of an analog scratch. RTN and MambaOFR both partially erase it, mistaking the railing for mechanical damage and blending it away with the aligned historical state, while DART's wider-receptive-field mask recognizes the railing as scene content and restores it intact.

\Cref{fig:false_positive_race} shows the complementary failure. The input frame contains a severe, unambiguous chemical crack running through the emulsion. The RTN and MambaOFR mask activations (top row) are almost uniformly flat: their unsupervised masks do not activate on this defect at all, so the restoration backbone receives no localization signal, and the crack survives essentially unchanged in both restored outputs (bottom row). DART's mask activates precisely along the crack, and the corresponding output attenuates it substantially. Taken together, the two figures show that, without a ground-truth target, an unsupervised mask can fail in either direction: activating on undamaged structure, or missing real, severe damage. Direct supervision keeps DART's localization anchored to the defect itself.

% --- mask false-positive on genuine scene structure (A Storm at Sea ROI) ---

% --- mask false-negative on real degradation (Race ROI) ---

\subsection{Ablation Study}
\label{sec:exp:ablation}
\Cref{tab:ablation} isolates the contribution of each core component by removing it from the full model (rows marked $-$). Every component earns its place: removing the Dilation Pyramid MaskNet, the recurrent mask pass, or the AdaLN conditioning each lowers every perceptual metric on real footage, with AdaLN conditioning the single most influential component, its removal alone drops MANIQA from $0.301$ to $0.263$. This pattern follows from how the components depend on one another: a sharper, supervised mask feeds a more informative signal into the Condition Encoder, so the severity vector $A(*)$ it produces genuinely reflects the damage present, and the restoration transformer is in turn modulated by a trustworthy signal rather than noise. Weaken the mask, and the conditioning that depends on it degrades with it.

% ---- Ablation table (leave-one-out from full DART) ----
\begin{table}[tbp]
    \centering
    \footnotesize
    \caption{Ablation study on the AbsoluteDegradation (real footage, no-reference) and Synthetic (full-reference) datasets. $-$ indicates removing the corresponding mechanism from the full DART model. Best scores are in \textbf{bold} and second-best scores are \underline{underlined}.}
    \label{tab:ablation}
    \setlength{\tabcolsep}{2pt}
    \renewcommand{\arraystretch}{1.05}
    \begin{tabularx}{\linewidth}{@{}
        >{\raggedright\arraybackslash\hsize=1.7\hsize}X|
        *{3}{>{\centering\arraybackslash\hsize=1.0\hsize}X}|
        *{3}{>{\centering\arraybackslash\hsize=0.77\hsize}X}@{}}
        \toprule
        & \multicolumn{3}{c}{\textbf{AbsoluteDegradation Dataset}} & \multicolumn{3}{c}{\textbf{Synthetic Dataset}} \\
        \cmidrule(lr){2-4}\cmidrule(lr){5-7}
        \textbf{Ablation} & CLIPIQA+~$\uparrow$ & MUSIQ~$\uparrow$ &
        MANIQA~$\uparrow$ & PSNR~$\uparrow$ & LPIPS~$\downarrow$ & SSIM~$\uparrow$ \\
        \midrule
        Input frames                  & 0.2483 & 26.842 & 0.1619 & 19.540 & 0.5421 & 0.5882 \\
        \midrule
        DART                          & \textbf{0.4334} & \textbf{51.230} & \textbf{0.3011} & 23.665 & \underline{0.1835} & 0.7294 \\
        - previous mask               & \underline{0.4332} & \underline{50.772} & \underline{0.2927} & \underline{23.815} & 0.1860 & \underline{0.7326} \\
        - AdaLN injection             & 0.4167 & 48.685 & 0.2629 & 23.622 & 0.1984 & 0.7295 \\
        - MaskNet                     & 0.4321 & 49.815 & 0.2905 & 23.733 & 0.1852 & 0.7309 \\
        - mask supervision            & 0.4173 & 49.451 & 0.2912 & \textbf{23.975} & \textbf{0.1787} & \textbf{0.7380} \\
        \bottomrule
    \end{tabularx}
\end{table}

The apparent counterexample, $-$\,mask supervision, is best on all full-reference proxy metrics, yet below DART on every real-footage metric. The same inversion holds across methods (supplementary Tab.~S1): ShiftNet and BasicVSR++, the two strongest on the proxy, rank fifth and last of seven in real-footage MUSIQ. A high proxy score can thus reflect fitting the synthetic degradations rather than generalizing to real film, so we keep the supervised configuration.

% ===============================================================
\section{Conclusion and Limitations}
\label{sec:conclusion}
We presented DART, a degradation-aware recurrent transformer that shifts archival film restoration from passive reconstruction to explicit damage-aware processing. DART predicts a directly supervised, continuous soft defect mask and propagates it through time. This mask guides recurrent temporal fusion and, via AdaLN-Zero conditioning, modulates the restoration backbone according to degradation severity. DART achieves state-of-the-art perceptual quality on two real-world benchmarks while adding just 1.3M trainable parameters, and its direct mask supervision prevents common unsupervised failure modes, preserving scratch-like scene structures and localizing severe damage.

However, DART currently targets monochrome footage and, like all flow-based propagators, relies heavily on optical flow accuracy, making extreme occlusions or highly persistent defects challenging. DART also has no internal shot-boundary handling, so across a cut the propagated state still carries the previous shot; cuts can be detected at inference time, \eg as in~\cite{absolutedegradation_eval_2026}, and the state reset without retraining. Defects more severe than the synthesis pipeline produces (\cref{fig:false_positive_race}) are localized but the attenuation may be limited. Furthermore, while we rely on established no-reference metrics, they do not fully substitute for human evaluation. Extending DART to color footage and validating it through formal perceptual studies remain promising future directions.

\bigskip\noindent
\begin{minipage}[c]{0.08\linewidth}\includegraphics[width=\linewidth]{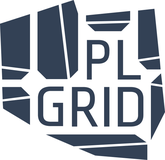}\end{minipage}\hfill
\begin{minipage}[c]{0.89\linewidth}
\textbf{Acknowledgements.} We gratefully acknowledge Polish high-performance computing infrastructure PLGrid (HPC Center: ACK Cyfronet AGH) for providing computer facilities and support within computational grant no. PLG/2025/018640.
\end{minipage}

% ---------------------------------------------------------------
\bibliographystyle{splncs04}
\bibliography{main}
\end{document}